\documentclass[letterpaper]{article}
\usepackage{aaai}
\usepackage{times}
\usepackage{helvet}
\usepackage{courier}
\nocopyright

\usepackage[usenames,dvipsnames]{xcolor}
\usepackage{amsmath,amsfonts,amssymb,amsthm}
\usepackage{url}
\usepackage{booktabs}

\newcommand{\citep}{\cite}
\newcommand{\citet}[1]{\citeauthor{#1}~\shortcite{#1}}

\makeatletter 
\newcommand\mynobreakpar{\par\nobreak\@afterheading} 
\makeatother

\theoremstyle{plain}
\newtheorem{theorem}{Theorem}

\newtheorem{corollary}[theorem]{Corollary}
\newtheorem{proposition}[theorem]{Proposition}

\newtheorem{property}[theorem]{Property}

\theoremstyle{definition}
\newtheorem{definition}{Definition}

\DeclareMathAlphabet{\mathbfsf}{\encodingdefault}{\sfdefault}{bx}{n}
\newcommand{\bm}[1]{\mathbf{#1}}

\newcommand{\plimpl}{\operatorname{:-}}

\def\xl{x}
\def\xls{\bm{x}}

\def\ys{\bm{y}}

\def\yl{y}
\def\yls{\bm{y}}

\def\zl{z}

\def\A{A}
\def\As{\bm{A}}

\def\B{B}

\def\w{w}

\newcommand{\false}{0}
\newcommand{\true}{1}

\def\P{P}

\def\zl{\mathsf{z}}
\def\Zp{\pred{Z}}

\def\Sp{\pred{S}}

\newcommand{\pred}[1]{\mathtt{#1}}
\newcommand{\s}{\smokes}
\newcommand{\smokes}{\pred{Smokes}}
\newcommand{\stress}{\pred{Stress}}

\newcommand{\f}{\friends}
\newcommand{\friends}{\pred{Friends}}
\newcommand{\mother}{\pred{MotherOf}}
\newcommand{\parent}{\pred{ParentOf}}
\newcommand{\female}{\pred{Female}}

\newcommand{\attends}{\pred{Attends}}

\newcommand{\toseries}{\pred{ToSeries}}
\newcommand{\series}{\pred{Series}}

\newcommand{\worksfor}{\pred{WorksFor}}
\newcommand{\boss}{\pred{Boss}}
\newcommand{\parents}{\pred{Parents}}
\newcommand{\first}{\pred{Adam}}

\newcommand{\alice}{\mathsf{A}}
\newcommand{\bob}{\mathsf{B}}
 
\newcommand{\Sk}{\mathsf{Sk}} 

\DeclareMathOperator*{\WFOMC}{WFOMC}
\DeclareMathOperator{\WMC}{WMC}

\DeclareMathOperator*{\predf}{pred}

\newcommand{\dom}{\mathbfsf{D}}

\newcommand{\wt}{\operatorname{w}}
\renewcommand{\wp}{\operatorname{w}}
\newcommand{\wf}{\operatorname{\bar{w}}}

\def\hyph{\mbox{-}}

\def\dDNNF{{\tt d\hyph{}DNNF}}
\def\SDD{{\tt SDD}}

\begin{document}

\title{Skolemization for Weighted First-Order Model Counting}

\author{Guy Van den Broeck\\
Computer Science Department\\
University of California, Los Angeles\\
\texttt{guyvdb@cs.ucla.edu}
\And
Wannes Meert\\  
Computer Science Department\\
KU Leuven \\
\texttt{wannes.meert@cs.kuleuven.be}
\And
Adnan Darwiche\\
Computer Science Department\\
University of California, Los Angeles\\
\texttt{darwiche@cs.ucla.edu}
}

\maketitle

\begin{abstract}
\begin{quote}
First-order model counting emerged recently as a novel reasoning task, at the core of efficient algorithms for probabilistic logics. We present a Skolemization algorithm for model counting problems that eliminates existential quantifiers from a first-order logic theory without changing its weighted model count. For certain subsets of first-order logic, lifted model counters were shown to run in time polynomial in the number of objects in the domain of discourse, where propositional model counters require exponential time. However, these guarantees apply only to Skolem normal form theories (i.e., no existential quantifiers) as the presence of existential quantifiers reduces lifted model counters to propositional ones. Since textbook Skolemization is not sound for model counting, these restrictions precluded efficient model counting for directed models, such as probabilistic logic programs, which rely on existential quantification. Our Skolemization procedure extends the applicability of first-order model counters to these representations. Moreover, it simplifies the design of lifted model counting algorithms.
\end{quote}
\end{abstract}

\section{Introduction}

Weighted model counting~(WMC) is a generalization of model counting~\citep{gomes2009model}. In model counting, also known as \#SAT,
one counts the number of satisfying assignments of a propositional sentence. 
In WMC, each assignment has an associated weight and the task is to compute the \emph{sum of the weights} of all satisfying assignments.
One application of WMC is to probabilistic graphical models. For example, 
exact inference algorithms for Bayesian networks encode probabilistic inference as a WMC task, which can then be solved by knowledge compilation~\citep{darwiche2002logical} or exhaustive DPLL search~\citep{sang2005solving}.

WMC also plays an important role in \emph{first-order probabilistic} representations. These became popular in recent years, in statistical relational learning~\citep{Getoor07:book} and probabilistic logic learning~\citep{DeRaedt2008-PILP}, which are concerned with modeling and learning complex logical and probabilistic interactions between large numbers of objects.
Efficient algorithms again reduce exact probabilistic inference to a WMC problem on a propositional knowledge base~\citep{Chavira2006,Fierens11,Fierens13}.
Encoding first-order probabilistic models into propositional logic retains a key advantage of the Bayesian network algorithms: WMC naturally exploits determinism and local structure in the probabilistic model~\citep{boutilier1996context,chavira2005compiling}.
A disadvantage is that the high-level first-order structure is lost.
\citet{Poole2003} observed that knowing the \emph{symmetries} that are abundant in first-order structure can speed up probabilistic inference.
Lifted inference algorithms reason about groups of objects as a whole, similar to the high-level reasoning of first-order resolution.
This has lead \citet{VdB11a} and \citet{gogatePTP} to propose \emph{weighted first-order model counting}~(WFOMC) as the core reasoning task underlying lifted inference algorithms. WFOMC assigns a weight to interpretations in finite-domain, function-free first-order logic, and computes the sum of the weights of all models.

Counting models at the first-order level has computational advantages. For certain classes of theories, knowing the first-order structure gives exponential speedups~\citep{VdBNIPS11}. 
For example, counting the models of a first-order universally quantified CNF with up to two logical variables per clause can always be done in time polynomial in the size of the domain of discourse. In contrast, a propositionalization of these CNFs will often have a treewidth polynomial in the domains size, and propositional model counting runs in exponential time.

One major limitation of first-order model counters, however, is that they require input in \emph{Skolem normal form} (i.e., without existential quantifiers). 
This is a common requirement for first-order automated reasoning algorithms, such as theorem provers. It is usually dealt with by Skolemization, 
which introduces Skolem constants and functions. However, the introduction of functions is problematic for first-order model counters as they expect a function-free input. 

The main contribution of this paper is a \emph{Skolemization procedure} that is specific for weighted first-order model counting. 
The procedure maps a logical input theory to an output theory that is devoid of existential quantifiers and functions, yet has an identical weighted first-order model count.
The procedure is modular, in that it remains sound when extending the input and output theories with a new sentence.
Furthermore, it is purely first-order as it is independent of the domain of discourse.

The proposed Skolemization algorithm has a range of implications.
First, it opens up new possibilities for lifted inference algorithms. For example,
on Markov Logic Networks with quantifiers~\citep{richardson2006markov}, and various forms of \emph{Probabilistic Logic Programs}~(e.g.,~\citet{de2007problog}), lifted algorithms currently provide little or no benefit over propositional ones. 
The main reason is that the WFOMC form of these representations generally contain existential quantifiers.
The proposed Skolemization algorithm allows us, 
for the first time, to perform lifted inference on these representations.
Second, there are \emph{liftability theorems} that define classes of theories for which WFOMC is domain-lifted, meaning that it runs in time polynomial in the domain size~\citep{JaegerStarAI12}.
These theorems had to assume Skolem normal form for the mentioned reason, but now apply more generally.
Finally, the Skolemization algorithm averts the need for special inference rules that deal with existential quantifiers, \emph{simplifying the design} of future WFOMC algorithms.

\section{Weighted First-Order Model Counting} \label{s:wfomc}

We start by formally defining the weighted first-order model counting task. 
We also compare it to propositional weighted model counting and discuss existing algorithms.

\subsection{Background}

Throughout this paper, we will work with the \emph{function-free} \emph{finite-domain} fragment of first-order logic~(FOL), which we now briefly review.
An {\em atom} $\pred{P}(t_1, \dots , t_n)$ consists of predicate $\pred{P}/n$ of arity $n$ followed by $n$ arguments, which are either \textit{constants} from a finite domain $\dom = \{\alice,\bob,\dots\}$ or \textit{logical variables} $\{\xl, \yl, \dots\}$.
We use $\yls$ to denote a sequence of logical variables.
A \emph{literal} is an atom or its negation.
A \emph{formula} combines atoms with logical connectives and quantifiers $\exists$ and $\forall$.
A logical variable $\xl$ is \emph{quantified} if it is enclosed by a $\forall \xl$ or $\exists \xl$. 
A \emph{free variable} is one that is not quantified.
A \emph{sentence} is a formula without free variables.
A formula is {\em ground} if it contains no logical variables.
A \emph{clause} is a disjunction of literals and a \emph{CNF} is a conjunction of clauses.
The \emph{groundings} of a quantifier-free formula is the set of formulas obtained by instantiating the free variables with any possible combination of constants from $\dom$.
The grounding of $\forall \xl,\phi$ and $\exists \xl,\phi$ is the conjunction resp.\ disjunction of all groundings of~$\phi$.

We will make use of \emph{Herbrand semantics}~\citep{hinrichs2006herbrand}, as is customary in statistical relational learning and probabilistic logic learning.
The \emph{Herbrand base} of sentence $\Delta$ for domain $\dom$ is the set of all ground atoms that can be constructed from predicates and constants in~$\dom$.
A \emph{Herbrand interpretation} is a truth-value assignment to all atoms in the Herbrand base. We will find it convenient to represent interpretations as sets of literals.
A Herbrand model of \(\Delta\) is a Herbrand interpretation $\omega$ that satisfies $\Delta$, denoted by~$\omega~\models_{\dom}~\Delta$.

\subsection{Definitions}

We first review propositional weighted model counting.

\begin{definition}[WMC]
  Given\mynobreakpar
  \begin{itemize}
    \item[--] a \emph{sentence} $\Delta$ in propositional logic over literals $\mathcal{L}$, and
    \item[--] a \emph{weight function} $\wp: \mathcal{L} \rightarrow \mathbb{R}^{\geq0}$,
  \end{itemize} 
  the \emph{weighted model count} (WMC) is
  \begin{align*}
    &\WMC(\Delta,\wp) = \sum_{\omega \models \Delta} \,\, \prod_{l \in \omega} \wp(l).
  \end{align*}
\end{definition}

\noindent WFOMC lifts WMC to the first-order level as follows.

\begin{definition}[WFOMC\footnote{This definition is based on \citet{VdB11a}. WFOMC is called \emph{Lifted WMC} in \citet{gogatePTP}.}]
  Given 
  \begin{itemize}
    \item[--] a \emph{sentence} $\Delta$ in FOL containing predicates $\mathcal{P}$,
    \item[--] a set of constants $\dom$, including the constants in $\Delta$, and
    \item[--] a pair of \emph{weight functions} $\wt,\wf: \mathcal{P} \rightarrow \mathbb{R}$,
  \end{itemize} 
  the \emph{weighted first-order model count} (WFOMC) is
  \begin{align*}
    &\WFOMC(\Delta,\dom,\wt,\wf) \\
    & \qquad\qquad = \sum_{\omega \models_{\dom} \Delta} \,\, \prod_{l \in \omega_\false} \wf(\predf(l)) \prod_{l \in \omega_\true} 
    \wt(\predf(l)),
  \end{align*}
  where $\omega_\false$ and $\omega_\true$ consists of the true, respectively false, literals in $\omega$, and $\predf$ maps literals to their predicate.
\end{definition}

The weight functions assign a weight to each predicate. The weight of a positive (negative) literal is the weight of its predicate in $\wt$ ($\wf$).
The weight of a model is the product of its literal weights.
Finally, the total count is the sum of the weights of all the Herbrand models of $\Delta$.

Our WFOMC definition deviates from WMC in two ways.
First, WMC directly assigns weights to \emph{literals}.
WFOMC instead assigns weights to \emph{predicates}, and defines literal weights in terms of predicate weights.
This distinguishes WFOMC from probabilistic databases~(see Section~\ref{s:rel}).
If for modeling reasons, certain literals need to be assigned unique weights, this can always be achieved by introducing new predicates.

Second, our definition permits predicate weights to be \emph{negative} numbers. 
Negative weights will turn out to be crucial for our Skolemization algorithm.
Historically, the WMC weight function has mostly been used to represent probabilities.
This led to the (sometimes implicit) assumption that weights are between zero and one, or at least non-negative.
Nevertheless, all exact weighted model counters we are aware of can handle negative weights.\footnote{In fact, the only underlying requirement of exact model counting approaches is that literal weights are elements from a \emph{commutative semiring}~\citep{kimmig2012algebraic}.} It appears that the positive weight assumption is more intrinsic to approximate weighted model counters~\citep{wei2005new,gogate2011samplesearch}.
Section~\ref{s:rel} discusses negative weights in more detail.

\subsection{Motivation}
\label{ss:motivation}

A WFOMC problem can always be propositionalized into a WMC problem. We can ground $\Delta$ for $\dom$, turn every atom in the Herbrand base into a propositional atom, and associate with every propositional literal the weight of its original predicate.
One may wonder why we define this task at the first-order level.

Our motivation is computational. Similar to how a single step of first-order resolution can perform a large number of propositional resolution steps, a WFOMC solver can often provide exponential speedups over WMC solvers.
First-order quantifiers make statements about groups of symmetric objects, which we can reason about jointly. 

Without going into algorithmic details, we will now illustrate this principle on concrete examples.
For the sake of simplicity, the examples are non-weighted model counting problems, corresponding to WFOMC problems where $\wt(\pred{P}) = \wf(\pred{P})=1$ for all predicates $\pred{P}$.
Consider $\Delta$ to be
\begin{align}  \label{f:ex1}
  \stress(\alice) \Rightarrow \smokes(\alice).
\end{align}
Assuming that $\dom=\{\alice\}$, every interpretation of $\stress(\alice)$ and $\smokes(\alice)$ satisfies $\Delta$, except when $\stress(\alice)$ is true and $\smokes(\alice)$ is false. Therefore, the model count is $3$.
Now let $\Delta$ be
\begin{align}  \label{f:ex2}
  \forall \xl,~\stress(\xl) \Rightarrow \smokes(\xl).
\end{align}
Without changing $\dom$, the model count is still $3$. 
When we expand $\dom$ to contain $n$ constants, we get $n$ independent copies of Formula~\ref{f:ex1}. For each person $\xl$, atoms $\stress(\xl)$ and $\smokes(\xl)$ can jointly take $3$ values, and the total model count becomes $3^n$.

This example already demonstrates the benefits of first-order counting.
A \emph{propositional} model counter on the groundings of Formula~\ref{f:ex2} would detect that all $n$ clauses are independent, recompute for every clause that it has $3$ models, and multiply these counts $n$ times. Propositional model counters have no elementary operation for exponentiation. 
A \emph{first-order} model counter reads from the first-order structure that it suffices to compute the model count of a single ground clause, and then knows to exponentiate. 
It never actually grounds the formula, and given the size of $\dom$, it runs in logarithmic time. This gives an exponential speedup over propositional counting, which runs in linear time.

These first-order counting techniques can interplay with propositional ones. Take for example $\Delta$ to be
\begin{align}  \label{f:ex3}
 \forall \yl,~\parent(\yl) \land \female \Rightarrow \mother(\yl).
\end{align}
This sentence is about a specific individual who may be a female, depending on whether the proposition $\female$ is true. 
We can separately count the models in which $\female$ is true, and those in which it is false (i.e., a Shannon decomposition). When $\female$ is false, $\Delta$ is satisfied, and the $\parent$ and $\mother$ atoms can take on any value. This gives $4^n$ models.
When $\female$ is true, $\Delta$ is structurally identical to Formula~\ref{f:ex2}, and has $3^n$ models. 
The total model count is then $3^n + 4^n$.

These concepts can be applied recursively to count more complicated formulas. Take for example
\begin{align*}
 \forall \xl, \forall \yl,~\parent(\xl,\yl) \land \female(\xl) \Rightarrow \mother(\xl,\yl).
\end{align*}
There is now a partition of the ground clauses into $n$ independent sets of $n$ clauses. The sets correspond to values of $\xl$, and the individual clauses to values of $\yl$.
The formula for each specific $\xl$, that is, each set of clauses, is structurally identical to Formula~\ref{f:ex3} and has count of $3^n + 4^n$.
The total model count is then $(3^n + 4^n)^n$.

The most impressive improvements are attained when propositional model counters run in time \emph{exponential} in $n$, 
yet first-order model counters run in \emph{polynomial} time. To consider an example where this comes up,  let $\Delta$ be
\begin{align}  \label{f:ex5}
 \forall \xl, \forall \yl,~\s(\xl) \land \f(\xl,\yl) \Rightarrow \s(\yl).
\end{align}
This time, the clauses in the grounding of $\Delta$ are no longer independent, and it would be wrong to simply exponentiate their counts.
Let us first assume that we know a partial interpretation of the $\smokes$ atoms with $k$ positive literals (i.e., $k$ people smoke).
The question is now: how many models extend this partial interpretation? Formula~\ref{f:ex5} encodes that a smoker cannot be friends with a non smoker. Hence, out of $n^2$ $\friends$ atoms, $k (n-k)$ have to be false, and the others can take either truth value. Thus, the number of models is $2^{n^2-k(n-k)}$.
Second, we know that there are $\binom{n}{k}$ partial interpretations with $k$ smokers, and $k$ can range from $0$ to $n$. This results in the total model count of
\begin{align*}
  \sum_{k=0}^n \binom{n}{k} 2^{n^2-k(n-k)}.
\end{align*}
In fact, the systems discussed in the next section can automatically construct this formula and compute the model count of Formula~\ref{f:ex5} in time polynomial in~$n$.
On the other hand, existing propositional WMC algorithms require time that is exponential in \(n\) on this problem. 
We note here that the \emph{treewidth} of the grounding of $\Delta$ is linear in~$n$. 

There are space considerations that motivate first-order model counting as well.
When converting a WFOMC problem to WMC, the grounding of $\Delta$ has size polynomial in the size of $\dom$, but the degree of this polynomial can be high.
When the grounding does not fit into memory, even approximate WMC becomes a problem.


\subsection{Algorithms}

Several algorithms exist for solving propositional WMC.
Exact solvers are based on either
exhaustive DPLL search~\citep{sang2005solving}, or knowledge compilation to a circuit language that supports efficient model counting, such as 
\dDNNF~\citep{darwiche2002logical,Chavira2008} or \SDD~\citep{ChoiKisaDarwiche13}.
Approximate WMC algorithms use local search~\citep{wei2005new} or sampling~\citep{gogate2011samplesearch}.

More recently, algorithms were introduced that directly solve the WFOMC task. They take a WFOMC problem and automatically generate and evaluate the types of expressions shown in the previous section. Their elementary operations include exponentiation, summation and binomial coefficients.
They are called \emph{lifted inference} algorithms.
In particular, two lifted algorithms were proposed for \emph{exact} WFOMC, one based on first-order knowledge compilation~\citep{VdB11a,VdBNIPS11,VdBThesis}, and the other based on first-order DPLL search~\citep{gogatePTP}.
Approximate algorithms were also proposed, including lifted importance sampling~\citep{gogatePTP,gogate2012advances}.
More generally, there is a large literature on exact and approximate \emph{lifted probabilistic inference} in statistical relational models,
which can be adapted to solve certain WFOMC tasks. See \citet{Kersting:2012} for an overview.

\subsection{Normal Forms}

It is common for logical reasoning algorithms to operate on \emph{normal form} representations instead of arbitrary sentences. 
For example, propositional SAT solvers and weighted model counters often expect CNF inputs.
We distinguish the following first-order normal forms.
\begin{itemize}
  \item[--] A theory in \emph{prenex normal form} consists of formulas
  $Q_1 \xl_1, \dots, Q_n \xl_n, ~\phi,$
where each $Q_i$ is either a universal or existential quantifier, and $\phi$ is quantifier-free. 
\item[--]  A theory in \emph{prenex clausal form} is a theory in prenex normal form where $\phi$ is a clause.
\item[--]  A theory in \emph{Skolem normal form} is a theory in prenex normal form where all $Q_i$ are universal quantifiers.
\item[--]  A \emph{first-order CNF} is a theory in Skolem and prenex clausal form. Thus, all sentences take the form
  $\forall \xl_1, \dots, \forall \xl_n, ~l_1 \lor \dots \lor l_m$.
\end{itemize}

Existing WFOMC algorithms require a theory to be in \emph{first-order CNF}.
The same requirement is often posed by automated theorem provers, such as first-order resolution.

\section{Skolemization for WFOMC}
\label{sec:skolemization}

It is well known that one can take any arbitrary formula and convert it to prenex clausal form.
This involves pushing negations inside, pushing quantifiers to the front, and distributing disjunctions over conjunctions.
The situation for Skolem normal form is different.

\subsection{Motivation}

Not every formula can be transformed into an equivalent Skolem normal form.
This problem is typically dealt with by \emph{Skolemization}, which eliminates existential quantifiers from a prenex normal form. 
This is done by replacing existentially quantified variables by Skolem constants and functions.
The result is not logically equivalent to the original formula, but only \emph{equisatisfiable} (i.e., satisfiable precisely when the original formula is satisfiable).

The standard Skolemization algorithm is specific to the satisfiability task and may be unsuitable for other tasks.
It is particularly unsuitable for WFOMC as it may produce a result with functions, which are not permitted in the WFOMC task. 
For example, standard Skolemization would transform the formula
\begin{align}  \label{f:boss}
 \forall \xl, \exists \yl,~\worksfor(\xl,\yl) \lor \boss(\xl)
\end{align}
into the following formula with the Skolem function $\Sk()$.
\begin{align*}
 \forall \xl,~\worksfor(\xl,\Sk(\xl)) \lor \boss(\xl).
\end{align*}
As soon as we allow functions, the Herbrand base becomes infinite, which makes the model counting task ill-defined, therefore,
ruling out standard Skolemization for WFOMC.\footnote{One could obtain a Skolem normal form by grounding existential quantifiers, replacing them by large, but finite disjunctions. While this may still permit limited runtime improvements on vacuous formulas, it is for all practical purposes equivalent to reducing the WFOMC problem to a WMC problem. Moreover, that transformation is dependent on the domain and leads to large formulas whose conversion to CNF blows up (e.g., when grounding $\exists \xl \forall \yl$).}

\subsection{Algorithm}

This section introduces a Skolemization technique for WFOMC.
It takes as input a triple $(\Delta,\wt,\wf)$ whose $\Delta$ is an arbitrary sentence and returns a triple $(\Delta',\wt',\wf')$  
whose $\Delta'$ is in Skolem normal form (i.e., no existential quantifiers).
Such a $\Delta'$ can then be turned into first-order CNF using standard transformations.
The proposed technique does not introduce functions. It satisfies two properties, one is essential and the other expands the applications of the technique.

The essential property is soundness.
\begin{property}[Soundness]
  \label{prop:soundness}
  Skolemization of $(\Delta,\wt,\wf)$ to $(\Delta',\wt',\wf')$ is sound iff for any $\dom$, we have that
  \begin{align*}
    \WFOMC(\Delta,\dom,\wt,\wf) = \WFOMC(\Delta',\dom,\wt',\wf').
  \end{align*}
\end{property}

To motivate the second property, we note that one may be interested in queries of the form $\WFOMC(\Delta \land \phi,\dom,\wt,\wf)$, 
where \(\Delta\), \(\wt\) and \(\wf\) are fixed, but where $\phi$ is changing. For example, we will see in Section~\ref{s:encodings} that 
probabilistic inference can be reduced to these types of queries. Therefore, we want to achieve a stronger form of soundness.
\begin{property}[Modularity]
  \label{prop:modularity}
  Skolemization of $(\Delta,\wt,\wf)$ to $(\Delta',\wt',\wf')$ is modular iff for any $\dom$ and any sentence $\phi$,
  \begin{align*}
    \WFOMC(\Delta \land \phi,\dom,\wt,\wf) = \WFOMC(\Delta' \land \phi,\dom,\wt',\wf').
  \end{align*}
\end{property}
That is, by replacing \(\phi\), one does not invalidate the Skolemization obtained under a different \(\phi\).

The proposed Skolemization algorithm eliminates existential quantifiers one by one.
Its basic building block is the following transformation.
\begin{definition} \label{def:elim-one}
  Suppose that $\Delta$ contains a subexpression of the form $\exists \xl, \phi(\xl,\yls)$, where $\phi(\xl,\yls)$ is an arbitrary sentence containing the free logical variables $\xl$ and $\yls$. Let $n$ be the number of variables in $\yls$.
  First, we introduce two new predicates: the {\em Tseitin predicate} $\Zp/n$ and the {\em Skolem predicate} $\Sp/n$. 
  Second, we replace the expression $\exists \xl, \phi(\xl,\yls)$ in $\Delta$ by the atom $\Zp(\yls)$, and append the formulas
  \begin{align*}
    \forall \yls, \forall \xl,~& \Zp(\yls) \lor \neg \phi(\xl,\yls)\\
    \forall \yls,~&\Sp(\yls) \lor \phantom{\neg}\Zp(\yls) \\
    \forall \yls,\forall \xl,~&\Sp(\yls) \lor \neg \phi(\xl,\yls).
  \end{align*}
  The functions $\wt'$ and $\wf'$ are equal to $\wt$ and $\wf$, except that $\wt'(\Zp) = \wf'(\Zp) = \wt'(\Sp) = 1$ and~$\wf'(\Sp) = -1$.
\end{definition}
In the resulting theory $\Delta'$, a single existential quantifier is now eliminated.
This building block can eliminate single universal quantifiers as well. When $\Delta$ contains a subexpression $\forall \xl, \phi(\xl,\yls)$, we replace it by $\neg \exists \xl, \neg \phi(\xl,\yls)$, whose existential quantifier can be eliminated with Definition~\ref{def:elim-one}.

We can now show the following.
\begin{theorem}[Modularity] \label{thm:modularity}
  Repeated application of Definition~\ref{def:elim-one} comprises a modular Skolemization algorithm.
\end{theorem}
\noindent The detailed proof can be found in the appendix.

\subsection{Intuition}

Our Skolemization algorithm implicitly tries to enforce an equivalence between the eliminated subexpression and the Tseitin predicate\footnote{This equivalence represents a set of propositional Tseitin encodings, in which each $\Zp$ atom is a Tseitin variable~\citep{tseitin1983complexity}.}, which is explicitly written as
\begin{align*}
\forall \yls,~ \Zp(\yls) \Leftrightarrow \left[\exists \xl, \phi(\xl,\yls)\right].
\end{align*}
This equivalence contains an existential quantifier so it cannot be represented explicitly.
Instead, the algorithms enforces a relaxed equivalence, represented by the three formulas in Definition~\ref{def:elim-one}.
The intuition is that by relaxing the equivalence we introduce additional models to the theory, but for every additional model with weight $W$, there is exactly one additional model with weight $-W$.\footnote{This is not dissimilar to the inclusion-exclusion principle.}
The WFOMC therefore stays the same.

The interaction between the three relaxed formulas, the intended equivalence, and the model weights becomes more apparent after a case analysis on $\Zp(\yls)$:
\begin{enumerate}
  \item 
    When $\Zp(\yls)$ is false, it implies that $\exists \xl,\phi(\xl,\yls)$ is false, which is intended.
    It also implies that $\Sp(\yls)$ is true, which does not change the model count, since we multiply by $1$.
  \item When $\Zp(\yls)$ is true, it implies that only three states of $\Sp(\yls)$ and $\exists \xl,\phi(\xl,\yls)$ are allowed:
    \begin{enumerate}
      \item $\exists \xl, \phi(\xl,\yls)$ is true and $\Sp(\yls)$ is true. This is again intended, because $\Zp(\yls)$ and $\exists \xl, \phi(\xl,\yls)$ are equivalent.
      \item $\exists \xl, \phi(\xl,\yls)$ is false and $\Sp(\yls)$ is true. This is an unintended state with a positive weight $W$.
      \item $\exists \xl, \phi(\xl,\yls)$ is true and $\Sp(\yls)$ is false. This is an unintended state with a weight $-W$. The negative weight comes from the fact that $\wf(\Sp)=-1$.
    \end{enumerate}
\end{enumerate}
The weights of the unintended models cancel each other out. 

\subsection{Examples}

We will now illustrate our Skolemization algorithm on concrete examples. Suppose that $\Delta$ is Formula~\ref{f:boss}, that is,
\begin{align*}
  \forall \xl, \exists \yl,~ \worksfor(\xl,\yl) \lor \boss(\xl).
\end{align*}
We can apply Definition~\ref{def:elim-one} to the subexpression $\exists \yl,~ \worksfor(\xl,\yl) \lor \boss(\xl)$, resulting in a $\Delta'$ equal to
\begin{align*}
  \forall \xl, ~& \Zp(\xl) \\[.3em]
  \forall \xl, \forall \yl,~& \Zp(\xl) \lor \neg \left[ \worksfor(\xl,\yl) \lor \boss(\xl) \right] \\
  \forall \xl, ~& \Zp(\xl) \lor \Sp(\xl) \\
  \forall \xl, \forall \yl,~& \Sp(\xl) \lor \neg \left[ \worksfor(\xl,\yl) \lor \boss(\xl) \right].
\end{align*}
The first formulas is the original formula with the subexpression substituted by $\Zp(\xl)$.

To get a better insight into the result, we will simplify it using first-order unit propagation~\citep{VdB11a} while noting that the first formula is a unit clause.
The simplified theory is
\begin{align*}
  \forall \xl, \forall \yl,~& \Sp(\xl) \lor \neg\worksfor(\xl,\yl) \\
  \forall \xl, ~& \Sp(\xl) \lor \neg\boss(\xl).
\end{align*}
We verify the correctness of this Skolemization as follows.
\begin{itemize}
  \item[--] When $\boss(\xl)$ is true, the formula is satisfied for $\xl$, and the models of $\Delta'$ are intended, that is, they correspond to models of $\Delta$. Indeed, $\Sp(\xl)$ is entailed to be true and the model weights are multiplied by one.

  \item[--] When $\boss(\xl)$ is false and  $\worksfor(\xl,\yl)$ is true for at least one $\yl$, then $\Sp(\xl)$ is entailed to be true. Again these models are intended, because $\exists \yl,~ \worksfor(\xl,\yl) \lor \boss(\xl)$ is now satisfied for $\xl$. The model weights are multiplied by one.

  \item[--] When $\boss(\xl)$ is false and $\worksfor(\xl,\yl)$ is false for all $\yl$ then $\Sp(\xl)$ can be either true or false. This is where unintended models appear, once with $\Sp(\xl)$ true and once with $\Sp(\xl)$ false. 
  Because they have opposing weights, the contributions of these unintended models cancel out.
\end{itemize}
As a second example, consider $\Delta$ to be
\begin{align*}
  \forall \xl, \exists \yl, \exists \zl,~ \parents(\xl,\yl,\zl) \lor \first(\xl).
\end{align*}
Skolemization of the inner existential quantifier results in
\begin{align*}
\forall \xl, \exists \yl,~& \Zp_1(\xl,\yl) \\[.3em]
\forall \xl, \forall \yl, \forall \zl,~& \Zp_1(\xl,\yl) \lor \neg \left[ \parents(\xl,\yl,\zl) \lor \first(\xl) \right] \\ 
\forall \xl, \forall \yl,~& \Zp_1(\xl,\yl) \lor \Sp_1(\xl,\yl) \\
\forall \xl, \forall \yl, \forall \zl,~& \Sp_1(\xl,\yl) \lor \neg \left[ \parents(\xl,\yl,\zl)  \lor \first(\xl) \right]
\end{align*}
This example shows the need for a Tseitin predicate~$\Zp_1$.
The first sentence still contains an existential quantifier. One more elimination and unit propagation step replaces that sentence by 
    $\forall \yl,\forall \xl,~\Sp_2(\yls) \lor \neg \Zp_1(\xl,\yl)$
and the result is in Skolem normal form.

\subsection{Properties}

Theorem~\ref{thm:modularity} suggests the repeated application of Definition~\ref{def:elim-one}, as long as the sentence contains an existential quantifier, or a universal quantifier not in prenex form.
This approach has one caveat: eliminating $\exists \xl, \phi(\xl,\yls)$ adds the expression $\neg \phi(\xl,\yls)$ to~$\Delta'$.
When we eliminate quantifiers from left to right, $\phi(\xl,\yls)$ itself can contain quantifiers.
This operation will introduce new quantifiers in $\neg \phi(\xl,\yls)$ and cause a blow up due to the duplication in the newly added formulas.
This can be avoided by eliminating from right to left, that is, from innermost to outermost.
We can show the following theorem, whose proof is in the appendix.
\begin{theorem}[Termination and Complexity]
  \label{thm:termination}
  Repeated application of Definition~\ref{def:elim-one} will terminate with a sentence in Skolem normal form.
  Moreover, this can be achieved in time polynomial in the size of $\Delta$.
\end{theorem}
\noindent
The resulting Skolem normal form sentence can subsequently be transformed into first-order CNF. When using Tseitin's  transformation~\citep{tseitin1983complexity}, this can even be done in polynomial time.

In our first example, the Tseitin predicate $\Zp$ could be removed from $\Delta'$ by unit propagation. The following proposition generalizes that observation.
\begin{proposition} \label{prop:easy-elim}
  Suppose that we are eliminating a subexpression $\exists \xl, \phi(\xl,\yls)$ from a sentence $\forall \yls, \exists \xls, \phi(\xl,\yls)$ using the procedure of Definition~\ref{def:elim-one}. That is, the existential quantifier in this subexpression is only preceded by universal quantifiers.
  Then, we can avoid adding Tseitin predicate $\Zp$ and instead define $\Delta'$ to be
  \begin{align*}
    \forall \yls,\forall \xl,~&\Sp(\yls) \lor \neg \phi(\xl,\yls).
  \end{align*}
\end{proposition}
\noindent
This simplifies the transformation when applicable, in particular when $\Delta$ is already in prenex normal form.

\section{WFOMC Encodings} \label{s:encodings}

We will show in this section how the proposed Skolemization technique can extend the scope of first-order model counters to new situations.
We will consider in particular one undirected first-order probabilistic language (Markov Logic) and one directed language (Probabilistic Logic Programs). 
First-order model counters currently apply to a subset of the first representation, and not to the second representation. With Skolemization, these
model counters can now be applied to both.
Our treatment is based on providing WFOMC encodings of these representations, to which our Skolemization technique is then applied.\footnote{These encodings are implemented in the WFOMC system: \url{http://dtai.cs.kuleuven.be/wfomc}}

Consider a first-order probabilistic model that induces the distribution~$\Pr\nolimits_\dom(.)$ for domain $\dom$.
A \emph{WFOMC encoding} of this model is a triple $(\Delta,\wt,\wf)$ which guarantees that for any sentence $\phi$ (usually a conjunction of literals) and domain $\dom$, we have that
\begin{align*}
  \Pr\nolimits_\dom(\phi) = \frac{\WFOMC(\Delta \land \phi, \dom, \wt, \wf)}{\WFOMC(\Delta, \dom, \wt, \wf)}.
\end{align*}

\subsection{Markov Logic Networks}

We will now introduce a WFOMC encoding for \emph{Markov logic networks}~(MLN)~\citep{richardson2006markov}.

\subsubsection{Representation}
An MLN is a set of tuples $(w,\psi)$, where $w$ is a real number representing a weight and $\psi$ is a formula in first-order logic.
When $w$ is infinite, $\psi$ represents a first-order logic constraint, also called a \textit{hard formula}.

Building further on the example given before, consider the following MLN
\begin{align}
1.3 \quad & \exists \yl,~ \worksfor(\xl,\yl) \lor \boss(\xl). \label{f:works}
\end{align}
This statement softens the logical sentence we saw earlier. Instead of saying that every person either has a boss, or is a boss, it states that worlds with many employed 
people are more likely. That is, it is now possible to have a world with unemployed people, but the more unemployed people there are, the lower the probability of that world.


The semantics of a first-order MLN \(\Phi\) is defined in terms of its \emph{grounding} for a given domain of constants $\dom$.
The grounding of $\Phi$ is the MLN obtained by first grounding all its quantifiers and then replacing each formula in \(\Phi\) with all its groundings (using the same weight).
With the domain $\dom = \{\alice,\bob\}$ (e.g., two people, Alice and Bob), the above first-order MLN represents the following grounding.
\begin{align*}
  1.3 \quad & \worksfor(\alice,\alice) \lor \worksfor(\alice,\bob) \lor \boss(\alice) \\
  1.3 \quad & \worksfor(\bob,\alice) \lor \worksfor(\bob,\bob) \lor \boss(\bob)
\end{align*}
This ground MLN contains six different random variables, which correspond to all groundings of 
atoms \(\worksfor(\xl,\yl)\) and \(\boss(\xl)\). This leads to a distribution over \(2^6\) possible worlds (i.e., interpretation). 
The weight
of each world is simply the product of all weights \(e^w\), where \((w,\gamma)\) is a ground MLN formula and \(\gamma\) is satisfied by the world. 
The weights of worlds that do not satisfy a hard formula are set to zero.
The probabilities of worlds are obtained by normalizing their weights.

\subsubsection{Encoding a Markov Logic Network}


\begin{definition} \label{def:mlnenc}
  The WFOMC encoding $(\Delta,\wt,\wf)$ of an MLN is constructed as follows.
  For each MLN formula $(w_i,\phi_i(\xls_i))$, where $\xls_i$ denotes the free logical variables in $\phi_i$, we introduce a parameter predicate $\pred{P_i}/|\xls_i|$.
  For each MLN formula, $\Delta$ contains the sentence $\forall \xls_i,~ \pred{P_{i}}(\xls_i) \Leftrightarrow \phi_i(\xls_i)$.
  The weight function sets $\wt(\pred{P_{i}}) = e^{w_i}$, $\wf(\pred{\P_{i}}) = 1$, and $\wt(\pred{Q}) = \wf(\pred{Q}) = 1$ for all other predicates $\pred{Q}$.
\end{definition}
\noindent
Each $\pred{P_i}$ captures the truth value of $\phi_i$ and carries its weight. Hard formulas can directly be encoded as constraints.

The encoding of Formula~\ref{f:works} has $\Delta$ equal to
\begin{align*}
  \forall \xl,~ \pred{P}(\xl) \Leftrightarrow ~& \exists \yl,~ \worksfor(\xl,\yl) \lor \boss(\xl).
\end{align*}
Its $\wt$ maps $\pred{P}$ to $e^{1.3}$ and all other predicates to $1$. Its $\wf$ maps all predicates to $1$.

As discussed in Section~\ref{s:wfomc}, WFOMC algorithms require first-order CNF input.
Definition~\ref{def:mlnenc} will only yield a $\Delta$ in Skolem normal form (and thus rewritable into CNF) if the MLN formulas are quantifier-free.
Then, the only quantifiers in $\Delta$ are the universal ones introduced by the encoding itself.
Therefore, \citet{VdB11a} and \citet{gogatePTP} resort to \emph{grounding all quantifiers} in the MLN formulas so as to obtain a CNF.
This makes the WFOMC encoding specific to the domain $\dom$, and partly removes first-order structure from the problem.

Our discussion is based on \citet{VdB11a}. It is similar to the encoding of \citet{gogatePTP}, whose parameter predicates have more arguments.
While these encodings are specific to MLNs, it is straightforward to generalize them to other undirected languages, such as \emph{parfactor graphs}~\citep{Poole2003}.


\subsubsection{Applying Skolemization}

We can now perform WFOMC inference in MLNs with quantifiers.
Skolemization and CNF conversion for the example above results in a $\Delta'$ equal to
\begingroup
\begin{align*}
  \forall \xl, ~& \pred{P}(\xl) \lor \neg\Zp(\xl) \\
  \forall \xl, ~& \neg\pred{P}(\xl) \lor \Zp(\xl) \\[.2em]
  \forall \xl, \forall \yl,~& \Sp(\xl) \lor \neg\worksfor(\yl,\xl) \\
  \forall \xl, ~& \Sp(\xl) \lor \neg\boss(\xl) \\
  \forall \xl, ~& \Sp(\xl) \lor \Zp(\xl) \\
  \forall \xl, \forall \yl,~ & \Zp(\xl) \lor \neg\worksfor(\xl,\yl) \\
  \forall \xl, \forall \yl,~ & \Zp(\xl) \lor \neg\boss(\xl)
\end{align*}%
\endgroup
This theory can be used for WFOMC inference.

\subsection{Probabilistic Logic Programs}
\label{ss:plp}

We now show a WFOMC encoding for a directed first-order probabilistic language.
The encoding is explained for the \emph{ProbLog} language \citep{de2007problog,Fierens13}.

\subsubsection{ProbLog Representation}

ProbLog extends logic programs with facts that are annotated with probabilities.
A ProbLog program $\Phi$ is a set of probabilistic facts $F$ and a regular logic program $L$. A probabilistic fact $p\!::\!a$ consists of a probability $p$ and an atom $a$. A logic program is a set of rules, with the form $\pred{Head}\plimpl\pred{Body}$, where the head is an atom and the body is a conjunction of literals.
For example,
\begin{align*}
& 0.1 :: \attends(\xl).  \\
& 0.3 :: \toseries(\xl) . \\
& \series \plimpl \attends(\xl) , \toseries(\xl).
\end{align*}
This program expresses that if more people attend a workshop, it more likely turns into a series of workshops.

The semantics of a ProbLog program $\Phi$ are defined by a distribution over the \emph{groundings} of the probabilistic facts for a given domain of constants $\dom$ \citep{Sato1995ICLP}.\footnote{Our treatment assumes a function-free and finite-domain fragment of ProbLog.
Starting from classical ProbLog semantics, one can obtain the a finite function-free domain for a given query by exhaustively executing the Prolog program and keeping track of the goals that are called during resolution.}
The probabilistic facts $p_i\!::\!a_i$ induce a set of possible worlds, one for each possible partition of $a_i$ in positive and negative literals. The set of true $a_i$ literals with the logic program $L$ define a well-founded model~\citep{vangelder1991well}.
The probability of such a model is the product of $p_i$ for all true $a_i$ literals and $1-p_i$ for all false $a_i$ literals.

For the domain $\dom = \{\A, \B\}$ (two people), the above first-order ProbLog program represents the following grounding:
\begin{align*}
& 0.1 :: \attends(\A). \\
& 0.1 :: \attends(\B). \\
& 0.3 :: \toseries(\A). \\
& 0.3 :: \toseries(\B). \\
& \series \plimpl \attends(\A) , \toseries(\A). \\
& \series \plimpl \attends(\B) , \toseries(\B).
\end{align*}
\noindent
This ground ProbLog program contains 4 probabilistic facts which corresponds to $2^4$ possible worlds. The weight of, for example, the world in which $\attends(\A)$ and $\toseries(\A)$ are true would be $0.1 \cdot (1-0.1) \cdot 0.3 \cdot (1-0.3) = 0.0189$ and the model would be $\{\attends(\A),\toseries(\A),\series\}$.


\subsubsection{Encoding a ProbLog Program}

The transformation from a ProbLog program to a first-order logic theory is based on Clark's completion  \citep{clark1978}. 
This is a transformation from logic programs to first-order logic.
For certain classes of programs, called \emph{tight} logic programs \citep{Fages1994}, it is correct, in the sense that every model of the logic program is a model of the completion, and vice versa.
Intuitively, for each predicate $\pred{P}$, the completion contains a single sentence encoding all its rules.
These rules have the form $\pred{P}(\xls) \plimpl b_i(\xls,\yls_i)$, where $b_i$ is a body and $\yls_i$ are the variables that appear in the body $b_i$ but not in the head.
The sentence encoding these rules in the completion is $\forall \xls,~\pred{P}(\xls) \Leftrightarrow \bigvee_i \exists \yls_i,~b_i(\xls,\yls_i)$.
If the program contains cyclic rules, the completion is not sound, and, it is necessary to first apply a conversion to remove positive loops~\citep{janhunen2004representing}.

\begin{definition} \label{def:plenc}
  The WFOMC encoding $(\Delta,\wt,\wf)$ of a tight ProbLog program has $\Delta$ equal to Clark's completion of $L$.
  For each probabilistic fact\footnote{If multiple probabilistic facts are defined for the same predicate, auxiliary predicates need to be introduced.} $p\!::\!a$ we set the weight function to $\wt(\predf(a)) = p$ and $\wf(\predf(a)) = 1-p$.
\end{definition}

Again, a Skolem normal form is required to use WFOMC. However, we get this form only when the variables that appear in the body of a rule also appear in the head of a rule.
This is not the case for most Prolog programs though. For example, if we apply Definition~\ref{def:plenc} to the example above, an existential quantifier appears in the sentence:
%
  \begin{align*}
    \series \Leftrightarrow \exists \xl, \attends(\xl) \land \toseries(\xl).
  \end{align*}
  Furthermore, $\wt$ maps $\attends$ to $0.1$ and $\toseries$ to $0.3$, and $\wf$ maps $\attends$ to $0.9$ and $\toseries$ to $0.7$. Both $\wt$ and $\wf$ are \(1\) for all other predicates.
This example is not in Skolem normal form and requires Skolemization before it can be processed by WFOMC algorithms.

\subsubsection{Applying Skolemization}



Skolemization followed by CNF conversion gives a $\Delta'$ equal to
\begin{align*}
  ~& \series \lor \neg\Zp \\
  ~& \neg\series \lor \Zp \\[.2em]
  \forall \xl,~& \Zp \lor \neg\attends(\xl) \lor \neg\toseries(\xl) \\
  ~& \Zp \lor \Sp \\
  \forall \xl,~& \Sp \lor \neg\attends(\xl) \lor \neg\toseries(\xl)
\end{align*}
Sentence $\Delta'$ is in Skolem normal form and is now processable by WFOMC algorithms.

A simple ProbLog program as the one above is identical to a noisy-or structure~\citep{Cozman2004ECAI}, popular in Bayesian network modeling.
Skolemization thus offers a fundamental method to lift first-order, directed structures, such as the noisy-or, in a generic manner (see also Section~\ref{s:rel}).

\section{Liftability Implications}

In our motivation for introducing first-order model counting, we touched upon the runtime and complexity improvements that can be attained by first-order counting.
These complexity improvements have inspired a particular notion of lifted inference, called \emph{domain-lifted} inference, which says that a WFOMC algorithm is lifted when it runs in time polynomial in the size of $\dom$~\citep{VdBNIPS11}.

While this notion of lifted inference may not capture everyone's perception of lifting, it does provide a clear formal framework.
In particular, we can now talk about classes of sentences $\Delta$ for which an algorithm is domain-lifted. We say that the algorithm is \emph{complete} for those classes.
We can also talk about classes of sentences $\Delta$ for which there exists, or cannot exist a domain-lifted algorithm. We call the former classes \emph{liftable}~\citep{JaegerStarAI12}.

All existing completeness and liftability theorems require that $\Delta$ is in first-order CNF. This requirement carries over from the existing WFOMC algorithms.
Given our Skolemization algorithm, we can now restate these theorems to apply more generally. For example, the positive liftability result of \citet{VdBNIPS11} becomes the following
\begin{corollary}
  Suppose that $\Delta$ is a theory of sentences with up to two logical variables, and otherwise arbitrary structure.
  The complexity of computing the WFOMC of $\Delta$ is polynomial in the size of $\dom$. That is, this class is domain-liftable.
\end{corollary}
Other notions of liftability also include queries $\phi$ in the complexity analysis, since they are important for lifted probabilistic inference. Based on \citet{van2012conditioning}, and \citet{VdBDarwiche13}, we can now claim the following.
\begin{corollary}
  Suppose that $\Delta$ is a theory of sentences with up to two logical variables, and otherwise arbitrary structure.
  The complexity of computing the WFOMC of $\Delta \land \phi$ is polynomial in the size of $\dom$ and~$\phi$, provided that $\phi$ is a conjunction of only unary literals, and binary literals of bounded Boolean rank.
\end{corollary}

These WFOMC liftability theorems have direct implications for all languages with a WFOMC encoding. For example, we can now say that MLNs with up to two logical variables per formula are domain-liftable, regardless of the quantifiers used. Previously, this was only true for quantifier-free MLNs.
We can now also show that ProbLog programs with up to two logical variables per clause are guaranteed to be liftable. This is the first such liftability result for probabilistic logic programs.

\section{Related Work} \label{s:rel}

In the encodings for MLNs and probabilistic logics, the weight functions (indirectly) represent probabilities and are therefore always positive.
Our Skolemization algorithm introduces negative weights. This might appear odd when interpreting the weights as negative probabilities. This issue has been discussed before. For example,  \citet{feynman1987negative} writes \emph{``Negative probabilities allow an abstract calculation which permits freedom to do mathematical calculations in any order simplifying the analysis enormously''}.

The potential of negative probabilities was already observed by \citet{Jha2012VLDBE} for answering queries in probabilistic databases and served as inspiration for our approach.
Probabilistic databases \citep{suciu2011probabilistic} are fundamentally a type of first-order probabilistic model.
It can be viewed as a special type of weighted model counting problem $(\Delta,\w)$, where the weight function encodes the probability $\w(t)$ with which a tuple $t$ can be found in the database.
A query on such a database is typically a union of conjunctive queries (UCQ), which corresponds to a monotone DNF sentence $\Delta$.
A noticeable difference with most WMC solvers (and WFOMC) is that the solvers for probabilistic databases expect the theory $\Delta$ to be in DNF instead of CNF.
Different from WFOMC is that although the query (i.e., $\Delta$) is first-order, the weight function is defined on the propositional level like in WMC.
Weights are thus assigned to ground literals (the tuples) whereas for WFOMC weights are assigned to predicates (the tables).
This allows WFOMC to exploit more types of symmetries.

\citet{Jha2012VLDBE} propose to extend probabilistic databases with MarkoViews, a representation similar to MLNs, in which each weighted formula is again a UCQ query, that is, a monotone DNF.
To compute the probability of a query,
they introduce negative tuple probabilities.

The use of negative probabilities has also come up for optimizing calculations for specific structures in probabilistic graphical models like noisy-or \citep{Diez2003IJIS}.
This particular case has been translated to the first-order case by \citet{Kisynski2009IJCAI} and resulted in an approach to lift noisy-or structures.
In Section~\ref{ss:plp} we showed how the application of Skolemization leads to lifting noisy-or and both methods turn out to output a similar encoding for this particular case.
Therefore, the approach followed by \citet{Kisynski2009IJCAI} can be considered a special case of the Skolemization algorithm applied to a noisy-or model.

\citet{jaeger2012lower} shows a negative liftability proof that uses \emph{relational Skolemization}. Similar to our approach, subexpressions containing an existential quantifier are transformed and relaxed to eliminate the quantifier. Relational Skolemization, however, does not guarantee a correct model count. It rather guarantees that if the weight of a model is non-zero it will also be non-zero in the Skolemized version.

\section{Conclusions}

In this paper, we introduced a Skolemization procedure that is sound for weighted first-order model counting.
It extends the applicability of first-order model counters to encodings which require an existential quantifier such as Markov logic models with quantifiers and probabilistic logic programs. It also extends the class of first-order sentences whose models we can count efficiently.

\section*{Acknowledgments}
This work was supported by ONR grant
\#N00014-12-1-0423, NSF grant \#IIS-1118122,
NSF grant \#IIS-0916161, and the Research Foundation-Flanders (FWO-Vlaanderen). GVdB is also at KU Leuven, Belgium.

\section*{Appendix}

\appendix

\section{Proof of Theorem~\ref{thm:modularity}}
\label{sec:proof}

We will now prove the sequence of steps that leads to the removal of an existential quantifier in $\Delta$ to obtain $\Delta'$, $\wt'$ and $\wf'$ while maintaining modularity. To replace the expression $\exists \xl, \phi(\xl,\yls)$ we perform the following steps.
\begin{description}
  \item[Isolate the Quantifier] 

  Introduce a new \emph{Tseitin} predicate $\Zp/n$, whose arity $n$ is the number of $\yls$ variables. Set $\wt'(\Zp) = \wf'(\Zp) = 1$ and for all other predicates $\pred{P}$, set $\wt'(\pred{P}) = \wt(\pred{P})$ and $\wf'(\pred{P}) = \wf(\pred{P})$.
  Construct $\Delta'$ by replacing the expression $\exists \xl, \phi(\xl,\yls)$ in $\Delta$ by the atom $\Zp(\yls)$, and appending the equivalence .

  In any grounding of $\Delta$, this step performs a Tseitin encoding of all groundings of $\exists \xl, \phi(\xl,\yls)$. The groundings of $\Zp(\yls)$ play the role of Tseitin variables. This step therefore satisfies Property~\ref{prop:modularity}.

  \item[Split the Equivalence] 

  Rewrite equivalence $\forall \yls,~ \Zp(\yls) \Leftrightarrow \exists \xl, \phi(\xl,\yls)$ as two implications, $\forall \yls,~ \Zp(\yls) \Rightarrow \exists \xl, \phi(\xl,\yls)$ and $\forall \yls,~ \Zp(\yls) \Leftarrow \exists \xl, \phi(\xl,\yls)$.
  In clausal form, these become
  \begin{align*}
    \forall \yls, \exists \xl,~& \neg \Zp(\yls) \lor  \phantom{\neg} \phi(\xl,\yls) \\
    \forall \yls, \forall \xl,~& \phantom{\neg}\Zp(\yls) \lor \neg \phi(\xl,\yls).
  \end{align*}

  This step satisfies Property~\ref{prop:modularity} because it is a logical equivalence.

  \item[Convert to a Feature] 

  Introduce a new \emph{Skolem predicate} predicate $\Sp/n$. Set $\wt(\Sp) = 1$ and $\wf(\Sp) = 0$ and replace the sentence 
   $ \forall \yls, \exists \xl,~ \neg \Zp(\yls) \lor \phi(\xl,\yls)$
  by
  \begin{align*}
   \forall \yls,~\Sp(\yls) \Leftrightarrow \exists \xl, \neg \Zp(\yls) \lor  \phi(\xl,\yls).
  \end{align*}

  In all models of the resulting theory where $\forall \yls, \exists \xl,~ \neg \Zp(\yls) \lor \phi(\xl,\yls)$ is not satisfied, there will exist a $\ys$ for which $\exists \xl, \neg \Zp(\yls) \lor  \phi(\xl,\yls)$ is not satisfied. This will cause at least one $\Sp(\yls)$ atom to be false in those models, which means that the weight of those models is multiplied by $0$. The weight of all other models remains the same.
  This step therefore satisfies Property~\ref{prop:modularity}.

  \item[Convert to an Implication]

  Set $\wf(\Sp) = -1$ and turn the equivalence $\forall \yls,~\Sp(\yls) \Leftrightarrow \exists \xl, \neg \Zp(\yls) \lor  \phi(\xl,\yls)$ into an implication $\forall \yls,~\Sp(\yls) \Leftarrow \exists \xl, \neg \Zp(\yls) \lor  \phi(\xl,\yls)$, which in clausal form becomes
  \begin{align*}
    \forall \yls,~&\Sp(\yls) \lor \phantom{\neg}\Zp(\yls) \\
    \forall \yls,\forall \xl,~&\Sp(\yls) \lor \neg \phi(\xl,\yls).
  \end{align*}


  Replacing the equivalence by an implication and changing $\wf(\Sp)$ to $-1$ is correct for the following reason.
  Let $\Sp(\yls) \Leftrightarrow \Sigma(\yls)$ be the above equivalence which is in $\Delta$, and let $\Gamma$ represent all other sentences in $\Delta$ (i.e., $\Delta \equiv (\Sigma(\yls) \Leftrightarrow \Sp(\yls)) \land \Gamma$).
  Our goal is now to construct a triple $(\Delta^\prime,\wt^\prime,\wf^\prime)$, where $\Delta^\prime \equiv (\Sigma(\yls) \Rightarrow \Sp(\yls)) \land \Gamma$, such that $\WFOMC(\Delta \land \phi,\dom,\wt,\wf) = \WFOMC(\Delta^\prime \land \phi,\dom,\wt^\prime,\wf^\prime)$ for all domains $\dom$ and all sentences $\phi$.
  
  Let $\Sigma(\alice)$ and $\Sp(\alice)$ be any arbitrary grounding of $\Sigma(\yls)$ and $\Sp(\yls)$.
  A case analysis on the values of $\Sigma(\alice)$ and $\Sp(\alice)$ shows that $\WFOMC(\Delta \land \phi,\dom,\wt,\wf)$ and $\WFOMC(\Delta^\prime \land \phi,\dom,\wt^\prime,\wf^\prime)$ consist of the following terms (for compactness we drop $\dom$ from the notation since it doesn't change).
  {\small
  \begin{center}
  \begin{tabular}{|cc|c|}
    \hline
    $\Sigma(\As)$\!\!\!\!\!\! & $\Sp(\As)$\!\!\! & $\WFOMC(\Delta \land \phi, \wt, \wf)$ \\
    \hline
    $\true$ & $\true$ & $\wt(\Sp) \cdot \WFOMC(\Gamma \land \Sigma(\As) \land \phi, \wt, \wf)$ \\
    $\true$ & $\false$ & 0 \\
    $\false$ & $\true$ & 0 \\
    $\false$ & $\false$ & $\wf(\Sp) \cdot \WFOMC(\Gamma \land \neg \Sigma(\As) \land \phi, \wt, \wf)$ \\
    \hline
  \end{tabular}\\
  \vspace{1em}
  \begin{tabular}{|cc|c|}
    \hline
    $\Sigma(\As)$\!\!\!\!\!\! & $\Sp(\As)$\!\!\!    & $\WFOMC(\Delta^\prime \land \phi,\wt^\prime, \wf^\prime)$                                            \\
    \hline
    $\true$   & $\true$  & $\wt^\prime(\Sp) \cdot \WFOMC(\Gamma \land \Sigma(\As)  \land \phi, \wt^\prime, \wf^\prime)$          \\
    $\true$   & $\false$ & 0                                                                                      \\
    $\false$  & $\true$  & $\wt^\prime(\Sp) \cdot \WFOMC(\Gamma \land \neg \Sigma(\As) \land \phi, \wt^\prime, \wf^\prime)$      \\
    $\false$  & $\false$ & $\wf^\prime(\Sp) \cdot \WFOMC(\Gamma \land \neg \Sigma(\As) \land \phi, \wt^\prime, \wf^\prime)$ \\
    \hline
  \end{tabular}
  \end{center}
  }
Note that $\wf(\Sp) = 0$ in the encoding of $\Delta$, and that thus
\begin{align*}
  & \WFOMC(\Delta \land \phi, \wt, \wf) =\\
  & \qquad \wt(\Sp) \cdot \WFOMC(\Gamma \land \Sigma(\As) \land \phi, \wt, \wf) \\
  & \qquad + 0 \cdot \WFOMC(\Gamma \land \neg \Sigma(\As) \land \phi, \wt, \wf) \\
  & \WFOMC(\Delta^\prime \land \phi, \wt^\prime, \wf^\prime) = \\
  & \qquad  \wt^\prime(\Sp) \cdot \WFOMC(\Gamma \land \Sigma(\As)  \land \phi, \wt^\prime, \wf^\prime) \\
  & \qquad + [\wt^\prime(\Sp) + \wf^\prime(\Sp)] \\
  & \qquad \qquad \cdot \WFOMC(\Gamma \land \neg \Sigma(\As) \land \phi, \wt^\prime, \wf^\prime)
\end{align*}
Setting $\wt^\prime(\pred{P}) = \wt(\pred{P})$ for all predicates $\pred{P}$ except for $\Sp$ ensures that $\WFOMC(\Gamma \land \Sigma(\As) \land \phi, \wt, \wf) = \WFOMC(\Gamma \land \Sigma(\As)  \land \phi, \wt^\prime, \wf^\prime)$,
that $\WFOMC(\Gamma \land \neg \Sigma(\As) \land \phi, \wt, \wf) = \WFOMC(\Gamma \land \neg \Sigma(\As) \land \phi, \wt^\prime, \wf^\prime)$. 
Furthermore, set $\wt(\Sp) = \wt^\prime(\Sp)$.
What remains for $\WFOMC(\Delta \land \phi, \wt, \wf)$ to equal $\WFOMC(\Delta^\prime \land \phi, \wt^\prime, \wf^\prime)$
is that $\wt^\prime(\Sp) + \wf^\prime(\Sp) = \wt(\Sp) + \wf^\prime(\Sp) = 0$, which is achieved by setting $\wf^\prime(\Sp) = -\wt(\Sp) = -1$.
  
\end{description}

\section{Proof of Theorem~\ref{thm:termination}}

\begin{proof}
  We begin by proving termination.
  Let the \emph{internal quantifier count} of a sentence be the number of quantifiers it contains, excluding the leading universal quantifiers.
  Suppose that a sentence has an internal quantifier count of $m$.
  We can select any subexpression that starts with a quantifier and apply Skolemization to it (potentially converting $\forall$ into $\exists$ first). This reduces the internal quantifier count of $\Delta$ to be at most $m-1$ because at least one quantifier is removed.
  New sentences are added, however, containing the Tseitin and Skolem predicates, and expressions  $\neg \phi(\xl,\yls)$.
  These sentences also have an internal quantifier count of at most $m-1$.
  Suppose that the sentences in $\Delta$ have an internal quantifier count of at most $m_{\mathit{max}}$.
  Applying Skolemization to one quantifier in each sentence reduces the maximal internal quantifier count to at most $m_{max}-1$.
  Therefore, by repeating this procedure for a finite number of steps, we obtain a theory with an internal quantifier count of zero, which is in Skolem normal form.
%
  
  Next, we prove polynomial complexity.
  We can remove the quantifiers in a sentence $\Delta$ one by one, starting from the innermost quantifier.
  The removed subexpression $\phi(\xl,\yls)$ does not contain any quantifiers, so the internal quantifier count of the added formulas is zero. They are in Skolem normal form.
  The innermost subexpression is replaced by a Tseitin predicate, reducing the internal quantifier count by one. 
  The number of required elimination steps before the entire sentence is in Skolem normal form is thus equal to the number of quantifiers in~$\Delta$.
  Moreover, the number of added formulas, and their size, is polynomial in the size of $\Delta$.
\end{proof}

\small
\bibliographystyle{aaai}
\bibliography{references.bib}

\end{document}